\documentclass{article}

\usepackage[numbers]{natbib}

\usepackage[final]{nips_2017}

\usepackage[utf8]{inputenc} 
\usepackage[T1]{fontenc}    
\usepackage{hyperref}       
\usepackage{url}            
\usepackage{booktabs}       
\usepackage{amsfonts}       
\usepackage{nicefrac}       
\usepackage{microtype}      

\usepackage[pdftex]{graphicx}
\usepackage{epstopdf}
\usepackage{mathtools}
\usepackage{array}
\usepackage{color}
\usepackage{bm}
\usepackage{multirow}
\usepackage{booktabs}
\usepackage{blkarray}
\usepackage{lineno} 
\usepackage{algorithm}
\usepackage{algorithmic}

\title{Deep Mutual Learning}

\author{
  Ying Zhang$^{1,2}$, Tao Xiang$^{2}$, Timothy M. Hospedales$^{3}$, Huchuan Lu $^{1}$\\
 $^1$Dalian University of Technology, China\\
$^2$Queen Mary University of London, UK\\
$^3$University of Edinburgh, UK\\
  \texttt{\{ying.zhang, t.xiang\}@qmul.ac.uk, t.hospedales@ed.ac.uk, lhchuan@dlut.edu.cn} \\
}

\begin{document}

\maketitle

\begin{abstract}
Model distillation is an effective and widely used technique to transfer knowledge from a teacher to a student network. The typical application is to transfer from a powerful large network or ensemble to a small network, that is better suited to low-memory or fast execution requirements. In this paper, we present a deep mutual learning (DML) strategy where, rather than one way transfer between a static pre-defined teacher and a student, an ensemble of students learn collaboratively and teach each other throughout the training process. Our experiments show that a variety of network architectures benefit from  mutual learning and achieve compelling results on CIFAR-100 recognition and Market-1501 person re-identification benchmarks. Surprisingly, it is revealed that no prior powerful teacher network is necessary -- mutual learning of a collection of simple student networks works, and moreover outperforms distillation from a more powerful yet static teacher. 
\end{abstract}

\section{Introduction}
 
Deep neural networks achieve state of the art performance on many problems, but are often very large in depth or width, and contain large numbers of parameters \cite{Resnet32,WideResnet16}. This has the drawback that they may be slow to execute or demand large memory to store, limiting their use in applications or platforms with low memory or fast execution requirements. This has led to a rapidly growing area of research on smaller and faster models. Achieving compact yet accurate models has been approached in a variety of ways including explicit frugal architecture design \cite{MobileNet17}, model compression \cite{han2016deepCompression}, pruning \cite{li2017pruninngConvNets}, binarisation \cite{rastegari2016xorNet} and most interestingly model distillation \cite{Distill15}. 

Distillation-based model compression relates to the observation \cite{bucila2006modelCompression,ba2014deepShallow} that small networks often have the same \emph{representation capacity} as large networks; but compared to large networks they are simply harder to train and find the right parameters that realise the desired function. That is, the limitation seems to lie in the difficulty of optimisation rather than in the network size \cite{ba2014deepShallow}. To better learn a small network, the  distillation approach starts with a powerful (deep and/or wide) teacher network (or network ensemble), and then trains a smaller student network to \emph{mimic} the teacher \cite{Distill15,ba2014deepShallow,parisotto2016transferRL,bucila2006modelCompression}. Mimicking the teacher's class probabilities \cite{Distill15} and/or feature representation \cite{ba2014deepShallow,FitNets15} conveys additional information beyond the conventional supervised learning target. The optimisation problem of learning to mimic the teacher turns out to be easier than learning the target function directly, and the much smaller student can  match or even outperform \cite{FitNets15} the larger teacher. 

In this paper we explore a different but related idea to model distillation -- that of \emph{mutual learning}. Distillation starts with a powerful  large and pre-trained teacher network and performs one-way knowledge transfer to a small untrained student. In contrast, in mutual learning we start with a pool of untrained students who learn simultaneously to solve the task together. Specifically, each student is trained with two losses: a conventional supervised learning loss, and a mimicry loss that aligns each student's class posterior with the class probabilities of other students. Trained in this way, it turns out that each student in such a peer-teaching based scenario learns significantly better than when learning alone in a conventional supervised learning scenario. Moreover student networks trained in this way achieve  better results than students trained by conventional distillation from a larger pre-trained teacher. Furthermore, while the conventional understanding of distillation requires a teacher larger and more powerful than the intended student, it turns out that in many cases mutual learning of several large networks also improves performance compared to independent learning.

It is perhaps not obvious why the proposed procedure should work at all. Where does the additional knowledge come from, when the learning process starts out with all small and untrained student networks? Why does it converge to a good solution rather than being hamstrung by groupthink as `the blind lead the blind'. Some intuition about these questions can be gained by considering the following: 
Each student is primarily directed by a conventional supervised learning loss, which means that their performance generally increases and they cannot drift arbitrarily into groupthink as a cohort. With supervised learning, all networks soon predict the same (true) labels for each training instance; but since each network starts from a different initial condition, their estimates of the probabilities of the next most likely classes vary. It is these secondary quantities that provide the extra information in distillation \cite{Distill15} as well as mutual learning.  \textcolor{black}{In mutual learning the student cohort effectively pools their collective estimate of the next most likely classes. Finding out -- and matching -- the other most likely classes for each training instance according to their peers  increases each student's posterior entropy \cite{chaudhar2017entropySGD,pereya2017outputDist}, which helps them to converge to a more robust (flatter) minima with better generalisation to testing data. This is related to very recent work on the robustness of high posterior entropy solutions (network parameter settings)  in deep learning  \cite{chaudhar2017entropySGD,pereya2017outputDist}, but with a much more informed choice of alternatives than blind entropy regularisation. }


Overall, mutual learning provides a simple but effective way to improve the generalisation ability of a  network by training collaboratively with a cohort of other networks. Compared with distillation by a pre-trained static large network, collaborative learning by small peers even achieves better performance.  Furthermore we observe that: (i) the efficacy increases with the number of networks in the cohort (by training on small networks only, more of them can fit on one GPU for effective mutual learning); (ii) it applies to a variety of network architectures, and to heterogeneous cohorts consisting of mixed big and small networks; and (iii) even large networks mutually trained in cohort improve performance compared to independent training.
Finally, we note that while our focus is on obtaining a single effective network, the entire cohort can also be used as a highly effective ensemble model.


 
{\bf Related Work}\quad The distillation-based approach to model compression has been proposed over a decade ago \cite{bucila2006modelCompression} but was recently re-popularised by \cite{Distill15}, where some additional intuition about why it works -- due to the additional supervision and regularisation of the higher entropy soft-targets -- was presented.  Initially, a common application was to distill the function approximated by a powerful model/ensemble teacher into a single neural network student \cite{bucila2006modelCompression,Distill15}. But later, the idea has been applied to distill powerful and easy-to-train large networks into small but harder-to-train networks \cite{FitNets15} that can even outperform their teacher.  Recently, distillation has been connected more systematically   to information learning theory \cite{paz2016distillationPriv} and SVM$+$ \cite{vapnik2015privilege} -- an intelligent teacher provides privileged information to the student. Here we address dispensing with the teacher altogether, and allowing an ensemble of students to teach each other in mutual distillation.

Other related ideas include Dual Learning \cite{DualLearning16} where two cross-lingual translation models teach each other interactively. But this only applies in this special translation problems where an unconditional within-language model is available to be used to evaluate the quality of the predictions, and ultimately provides the supervision that drives the learning process. In contrast, our mutual learning approach applies to general classification problems. While conventional wisdom about ensembles prioritises diversity \cite{kuncheva2003diverseEnsemble}, our mutual learning approach reduces diversity in the sense that all students become somewhat more similar by learning to mimic each other. However, our goal is not necessarily to produce a diverse ensemble, but to enable networks to find robust solutions that generalise well to testing data, which would otherwise be hard to find through conventional supervised learning.

\section{Deep Mutual Learning}


\begin{figure}[t]
  \centering
  \includegraphics[width=0.8\linewidth]{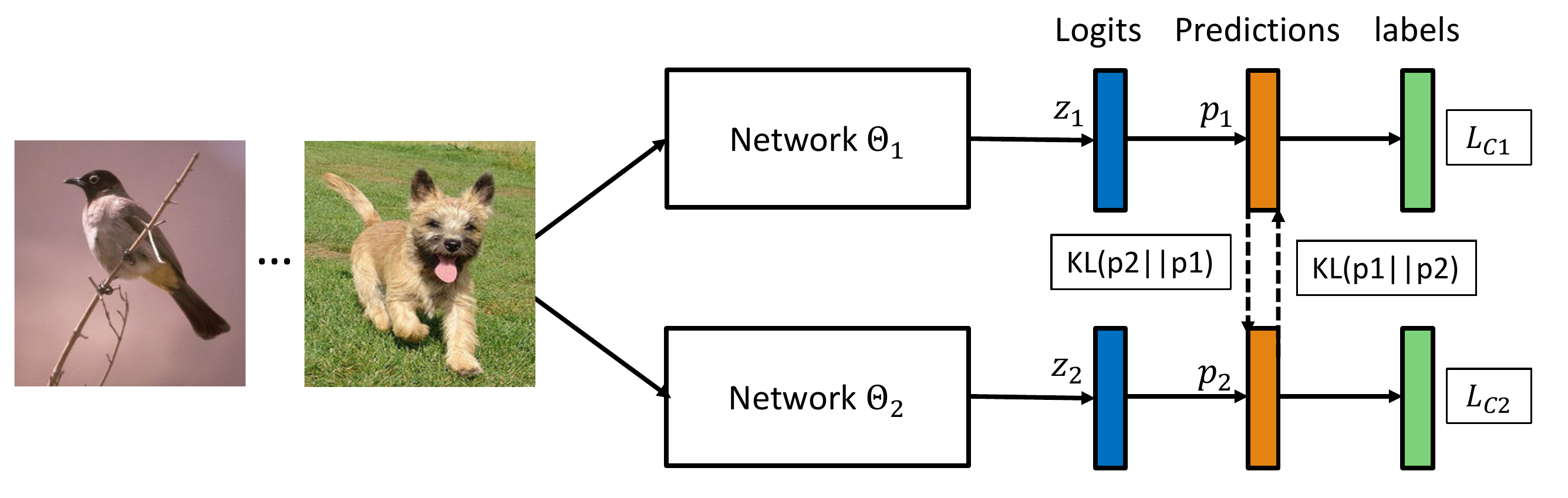}
  \caption{Deep Mutual Learning (DML) schematic. Each network is trained with a supervised learning loss, and a KLD-based mimcry loss to match the probability estimates of its peers.}\label{fig:DML}
\end{figure}

\subsection{Formulation}
We formulate the proposed DML approach with a cohort of two networks (see Fig.~\ref{fig:DML}). Extension to more networks is straightforward (see Sec.~\ref{sec:extension-to-multi-net}).
Given $N$ samples $\mathcal{X}=\{\boldsymbol{x}_{i}\}_{i=1}^{N}$ from $M$ classes, we denote the corresponding label set as $\mathcal{Y}=\{{y}_{i}\}_{i=1}^{N}$ with $y_{i} \in \{1, 2, ..., M \} $.  
The probability of class $m$ for sample $\boldsymbol{x}_i$ given by a neural network $\Theta_{1}$  is computed as
 \begin{equation}
p_{1}^{m}(\boldsymbol{x}_i) = \frac{exp(z_{1}^{m})}{\sum_{m=1}^{M} exp(z_{1}^{m})}, 
 \label{eq:pred}
 \end{equation}
where the logit $z^{m}$ is the output of  the ``softmax'' layer in $\Theta_{1}$.

For multi-class classification, the objective function to train the network $\Theta_{1}$   is defined as the cross entropy error between the predicted values and the correct labels,  
 \begin{equation}
L_{C_{1}} = - \sum_{i=1}^{N} \sum_{m=1}^{M} I(y_{i}, m) \log(p_{1}^{m}(\boldsymbol{x}_i))  ,
 \end{equation}
 with an indicator function $I$ defined as $I(y_{i}, m) =\left\{\begin{array}{ll} 1 \quad & {y_{i} = m} \\
0 \quad & {y_{i}\neq m} \end{array}\right.$.

The conventional supervised loss trains the network to predict the correct labels for the training instances. To improve the generalisation performance of $\Theta_{1}$ on testing instances, we use another peer network $\Theta_{2}$ to provide training experience in the form of its posterior probability $\bm{p}_2$. To measure the match of the two network's predictions $\bm{p}_1$ and $\bm{p}_2$, we adopt the Kullback Leibler (KL) Divergence.

 The KL distance from $\boldsymbol{p}_{1}$ to $\boldsymbol{p}_{2}$  is computed as
 \begin{equation}
 D_{KL}(\boldsymbol{p}_{2}\|\boldsymbol{p}_{1}) = \sum_{i=1}^{N} \sum_{m=1}^{M} p_{2}^{m}(\boldsymbol{x}_i) \log \frac{p_{2}^{m}(\boldsymbol{x}_i)}{p_{1}^{m}(\boldsymbol{x}_i)}.
  \end{equation}
 
The  overall loss function  $L_{\Theta_{1}}$ for network $\Theta_{1}$ is defined as 
  \begin{equation}
   \label{eq:2DML}
L_{\Theta_{1}} = L_{C_{1}} + D_{KL}(\boldsymbol{p}_{2}\|\boldsymbol{p}_{1}).
 \end{equation}

Similarly, the objective loss function $L_{\Theta_{2}}$ for network $\Theta_{2}$ can be computed as 
  \begin{equation}
L_{\Theta_{2}} = L_{C_{2}} + D_{KL}(\boldsymbol{p}_{1}\|\boldsymbol{p}_{2}).
 \end{equation}
 
In this way each network learns both to correctly predict the true label of training instances (supervised loss $L_C$) as well as to match the probability estimate of its peer (KL mimicry loss).
 



\subsection{Optimisation}
The mutual learning strategy is performed in each mini-batch based model update step and throughout the whole training process. At each iteration, we compute the predictions of the two models and update both networks' parameters according to the predictions of the other. The optimisation of $\Theta_{1}$ and $\Theta_{2}$ is conducted iteratively until convergence.  The  optimisation details  are summarised in Algorithm~1.

\begin{table}[!h]
\begin{tabular}{p{.99\linewidth}}
\toprule
\textbf{Algorithm 1:  Deep Mutual Learning  }\\
\midrule
\textbf{\emph{Input: }} Training set $\mathcal{X}$, label set $\mathcal{Y}$, learning rate $\gamma_{1, t}$ and  $\gamma_{2, t}$ . \\ 
\textbf{\emph{Initialize: }} Models $\Theta_{1}$ and $\Theta_{2}$ to different initial conditions.  \\

\textbf{\emph{Repeat :}} \\
\quad  $t = t + 1$ \\
\quad  Randomly sample data $\boldsymbol{x}$ from $\mathcal{X}$. \\

\quad  \textbf{1: } Update the predictions $\boldsymbol{p}_{1}$ and $\boldsymbol{p}_{2}$  of 
$\boldsymbol{x}$ by \eqref{eq:pred} for  the current mini-batch \\

\quad  \textbf{2: } Compute the stochastic gradient and update $\Theta_{1}$ :   \\

  \begin{equation}
\Theta_{1} \leftarrow \Theta_{1} + \gamma_{1,t} \frac{\partial L_{\Theta_{1}}}{\partial \Theta_{1} }
 \end{equation}
\quad   \textbf{3: } Update the predictions $ \boldsymbol{p}_{1}$  and  $ \boldsymbol{p}_{2}$  of 
$\boldsymbol{x}$ by  \eqref{eq:pred}  for the current mini-batch \\
\quad   \textbf{4: } Compute the stochastic gradient and update $\Theta_{2}$ :     \\
  \begin{equation}
\Theta_{2} \leftarrow \Theta_{2} + \gamma_{2,t} \frac{\partial L_{\Theta_{2}}}{\partial \Theta_{2} }
 \end{equation}
\textbf{\emph{Until :}}  convergence \\

\bottomrule
\end{tabular}
\vspace{-2mm}
\nonumber
\end{table}

\subsection{Extension to Larger Student Cohorts}
\label{sec:extension-to-multi-net}

The proposed DML approach naturally extends to more networks in the student cohort.
Given $K$ networks $\Theta_{1}, \Theta_{2}, ..., \Theta_{K} (K\geq 2)$, the objective function for optimising $\Theta_{k}, (1 \leq k \leq K)$ becomes
  \begin{equation}
  \label{eq:MDML}
L_{\Theta_{k}} = L_{C_{k}} + \frac{1}{K-1}\sum_{l=1, l \neq k}^{K} D_{KL}(\boldsymbol{p}_{l}\|\boldsymbol{p}_{k}).
 \end{equation}
Equation~\eqref{eq:MDML} indicates that with $K$ networks, DML for each student effectively takes  the other $K-1$ networks in the cohort as $K-1$ teachers to provide learning experience. Equation~\eqref{eq:2DML} is now a special case of \eqref{eq:MDML} with $K=2$. Note that we have added the coefficient $\frac{1}{K-1}$ to make sure that  the training is mainly directed by supervised learning of the true labels. The optimisation for DML with more than two networks is a straightforward extension of Algorithm 1. It can be distributed by learning each network on one device and passing the small probability vectors between devices.

{\color{black} With more than two networks, an  interesting alternative learning strategy for DML is to take the ensemble of all the other $K-1$  networks as a single teacher to provide an averaged learning experience, which would be very similar to the distillation approach but performed at each mini-batch model update. Then the objective function of $\Theta_{k}$ can be written as 
  \begin{equation}
  \label{eq:EMDML}
L_{\Theta_{k}} = L_{C_{k}} + D_{KL}(\boldsymbol{p}_{avg}\|\boldsymbol{p}_{k}), \quad  \boldsymbol{p}_{avg} = \frac{1}{K-1}\sum_{l=1, l \neq k}^{K} \boldsymbol{p}_{l}.
 \end{equation}
In our experiments (see Sec.~\ref{sec:why DML works}), we find that this DML strategy  with a single ensemble teacher or DML\_e leads to worse performance than DML with $K-1$ teachers. This is because the model averaging step (Equation (\ref{eq:EMDML})) to build the teacher ensemble makes the teacher's posterior probabilities more peaked at the true class, thus reducing the posterior entropy over all classes. It is therefore contradictory to one of the objectives of DML which is to produce robust solutions with high posterior entropy.  }


\section{Experiments}

\subsection{Datasets and Settings}

{\bf Datasets}\quad Two  datasets are used in our experiments.  
The {\bf CIFAR-100}~\cite{cifar} dataset consists of $32 \times 32$ color images drawn from 100 classes, which are split into 50,000 train and 10,000 test images.  The Top-1 classification accuracy is reported.
The {\bf Market-1501}~\cite{market1501} dataset is widely used in the person re-identification problem which aims to associate people across different non-overlapping camera views. It contains 32,668 images of 1,501 identities captured from six camera views, with 751 identities for training and 750 identities for testing. As per state of the art approaches to this problem \cite{Rerank16}, we train the network for 751-way classification and use the resulting feature \textcolor{black}{of the last pooling layer} as a representation for \textcolor{black}{nearest neighbour} matching at testing. This is a more challenging dataset than CIFAR-100 because the task is instance recognition thus more fine-grained, and the dataset is smaller with more classes. For evaluation, the standard Cumulative Matching Characteristic (CMC) Rank-k accuracy and mean average precision (mAP) metrics \cite{market1501}  are used.

{\bf Implementation Details}\quad We implement all networks and training procedures in TensorFlow \cite{DBLP:journals/corr/AbadiABBCCCDDDG16} and conduct all experiments on an NVIDIA GeForce GTX 1080 GPU. 
For CIFAR-100, we follow the experimental settings of \cite{WideResnet16}. Specifically, we use SGD with Nesterov momentum and set the initial learning rate to 0.1, momentum to 0.9 and mini-batch size to 64. The learning rate dropped by 0.1 every 60 epochs and we train for 200 epochs. The data augmentation includes horizontal flips and random crops from image padded by 4 pixels on each side, filling missing pixels with reflections of original image. 
For Market-1501, we use Adam optimiser~\cite{Adam15}, with learning rate $lr=0.0002$, $\beta_{1}=0.5$,  $\beta_{2}=0.999$ and a mini-batch size  of 16. We train all the models for 100,000 iterations. We also report  results with and without pre-training on ImageNet.


{\bf Model Size}\quad The networks used in our experiments includes compact networks of typical student size: Resnet-32~\cite{Resnet32} and  MobileNet \cite{MobileNet17}; as well as large networks of typical teacher size: InceptionV1 \cite{InceptionV1} and  Wide ResNet WRN-28-10 \cite{WideResnet16}. Table \ref{tabnets} compares the number of parameters of all the networks on  CIFAR-100. 

\begin{table}[h!]
\begin{center}
\begin{tabular}{ccccc}
\hline
\: 		& 		ResNet-32		  		 & 		MobileNet 		 & 	InceptionV1 & 		WRN-28-10\\
\hline
\# parameters  &	 0.5M  		 & 	3.3M 		& 			7.8M   	& 		36.5M	\\
\hline
\end{tabular}
\end{center}
\caption{Number of parameters on the CIFAR-100 dataset}\label{tabnets}
\end{table}

\subsection{Results on CIFAR-100}
Table~\ref{tabcifar} compares the Top-1 accuracy of the CIFAR-100 dataset obtained by various architectures in a two-network DML cohort. From the table we can make the following observations: (i) All the different network combinations among ResNet-32, MobileNet and WRN-28-10 improve performance when learning in a cohort compared to learning independently, indicated by the all positive values in the ``DML-Independent''  columns. (ii) The networks with smaller capacity (ResNet-32 and MobileNet) generally benefit more from DML. (iii) Although WRN-28-10 is a much larger network than MobileNet or ResNet-32 (Table~\ref{tabnets}) it  still benefits from being trained together with a smaller peer. (iv) Training a cohort of large networks (WRN-28-10) is still beneficial compared to learning alone. Thus in contrast to the conventional wisdom of model distillation, we see that a large pre-trained teacher is not necessary to obtain benefits, and multiple large networks can still benefit from our distillation-like process. 

\begin{table}[h!]
\begin{center}
\begin{tabular}{cccccccc}
\hline
\multicolumn{2}{c}{Network Types}  & \multicolumn{2}{c}{\textcolor {black}{Independent }} & \multicolumn{2}{c}{DML } & \multicolumn{2}{c}{\textcolor{black}{DML-Independent}} \\
Net  1 & Net  2 & Net 1 & Net 2 & Net 1 & Net 2 & Net 1 & Net 2\\
\hline
Resnet-32           &   Resnet-32    &	68.99    &   68.99   &   	71.19  	&  70.75  &  1.20   &  1.76	\\
WRN-28-10 		 &   Resnet-32    &		78.69    &   68.99 	&   	78.96  	&   70.73  &  0.27  &  1.74  \\
MobileNet 		 &   Resnet-32    &    73.65    &   68.99 	&   	76.13  	&   71.10  &  2.48  &  2.11 \\
MobileNet        &   MobileNet     &   73.65    &   73.65 	&   	76.21  	&   76.10  &  2.56  & 2.45 \\
WRN-28-10      &   MobileNet    &    78.69   &   73.65    	&   	80.28 	&   77.39  & 1.59  & 3.74 \\
WRN-28-10      &   WRN-28-10   &   78.69    &   78.69 	&   	80.28  	&   80.08  & 1.59 & 1.39 \\
\hline
\end{tabular}
\end{center}
\caption{Top-1 accuracy (\%) on the CIFAR-100 dataset. ``DML-Independent'' measures the difference in accuracy between the network learned with DML and the same network learned independently. }\label{tabcifar}
\end{table}


\subsection{Results on Market-1501}
Table \ref{tabcompmarket} summarises the mAP (\%) and rank-1 accuracy (\%) of Market-1501  with/without DML, as well as the comparison against existing state of the art methods. Each MobileNet is trained in a two-network cohort and the averaged performance of the two networks in the cohort is reported. We  can see that DML greatly improves the performance of MobileNet compared to independent learning, both with and without pre-training on ImageNet. It can also be seen that the performance of the proposed DML approach trained with two MobileNets significantly outperforms prior state-of-the-art deep learning methods.

\begin{table}[h!]
\begin{center}
\begin{tabular}{cccccc}
\hline
\multirow{2}{*}{Method}  &{ImageNet} & \multicolumn{2}{c}{Single-Query} &   \multicolumn{2}{c}{Multi-Query} \\
   \:                                          & Pretrain? &     mAP     &      Rank-1    &     mAP    &     Rank-1     \\
\hline
CAN~\cite{CAN16}                    & yes   &   24.40   &    48.20    &       -           &     -       \\
Gated S-CNN~\cite{GSCNN16}   & no  &   39.55   &   65.88    &    48.45      &    76.04  \\
Siamese LSTM~\cite{SLSTM16}  & no   &      -        &        -       &    35.30      &   61.60   \\
$k$-reciprocal Re-ranking~\cite{Rerank16}                          & yes   & 63.63     &   77.11  &     -   &  -  \\
\hline
MobileNet  &  no   & 46.07 & 72.18 & 54.31  & 79.81 \\
MobileNet+DML  & no  & 52.15 & 76.90 & 60.97  & 83.48 \\
MobileNet & yes & 60.68 & 83.94 & 68.25  & 87.89\\
MobileNet+DML  & yes  & \textbf{68.83} & \textbf{87.73} & \textbf{77.14}  & \textbf{91.66} \\
\hline
\end{tabular}
\end{center}
\caption{Comparative results on the Market-1501 dataset }\label{tabcompmarket}
\end{table}

\subsection{Comparison with Distillation}
As our method is strongly related to model distillation, we next provide a focused comparison to Distillation~\cite{Distill15}. 
Table~\ref{tabmarket} compares our DML with model distillation  where the teacher network (Net 1) is pre-trained and provides fixed posterior targets for the student network (Net 2). As expected the conventional distillation approach from a powerful pre-trained teacher does indeed improve the student performance compared to independently learning the student (1 distills 2 versus Net 2 Independent). However, the results show that not only is a pre-trained teacher unnecessary; but training both networks together in deep mutual learning provides a clear improvement compared to distillation (1 distills 2 versus DML Net 2). This implies that in the process of mutual learning the network that would play the role of teacher actually becomes better than a pre-trained teacher, via learning from interactions with an a-priori untrained  student. Finally, we note that on Market-1501 training two compact MobileNets together provides a similar  boost over independent learning compared to mutual learning with InceptionV1 and MobileNet: Peer teaching of small networks can be highly effective. In contrast, using the same network as teacher in model distillation actually makes the student worse than independent learning (the last row 1 distills 2 (45.16) vs.~ Net 2 Independent (46.07)). 


\begin{table}[h!]
\begin{center}
\begin{tabular}{cccccccc}
\hline
\multirow{2}{*}{Dataset} & \multicolumn{2}{c}{Network Types}  & \multicolumn{2}{c}{\textcolor {black}{Independent }} & 1 distills 2  & \multicolumn{2}{c}{DML }  \\
 \: & Net1 & Net 2 & Net 1 & Net 2 & Net 2 & Net 1 & Net 2 \\
\hline
\multirow{2}{*}{CIFAR-100}  &  WRN-28-10   &  ResNet-32    & 78.69  & 68.99  & 	 69.48  &  78.96    & 	  70.73   \\
\: & MobilNet  &  ResNet-32  & 73.65	&  68.99  &  69.12  &	  76.13  &   71.10   \\
\hline 
\multirow{2}{*}{Market-1501} & Inception V1  &  MobileNet   & 65.26	& 46.07  &  49.11 &	  65.34  &   52.87   \\
\: &  MobileNet    &  MobileNet    & 46.07  & 46.07  & 	 45.16  &  52.95    & 	  51.26   \\

\hline 

\end{tabular}
\end{center}
\caption{Comparison with distillation on CIFAR-100 (Top-1 accuracy (\%)) and Market-1501 dataset (mAP (\%))}\label{tabmarket}
\end{table}

\subsection{DML with Larger Student Cohorts}
\begin{figure}[htpb]
\centering
\begin{tabular}{@{}c@{}c}
  \includegraphics[width=0.47\linewidth, height=0.34\linewidth]{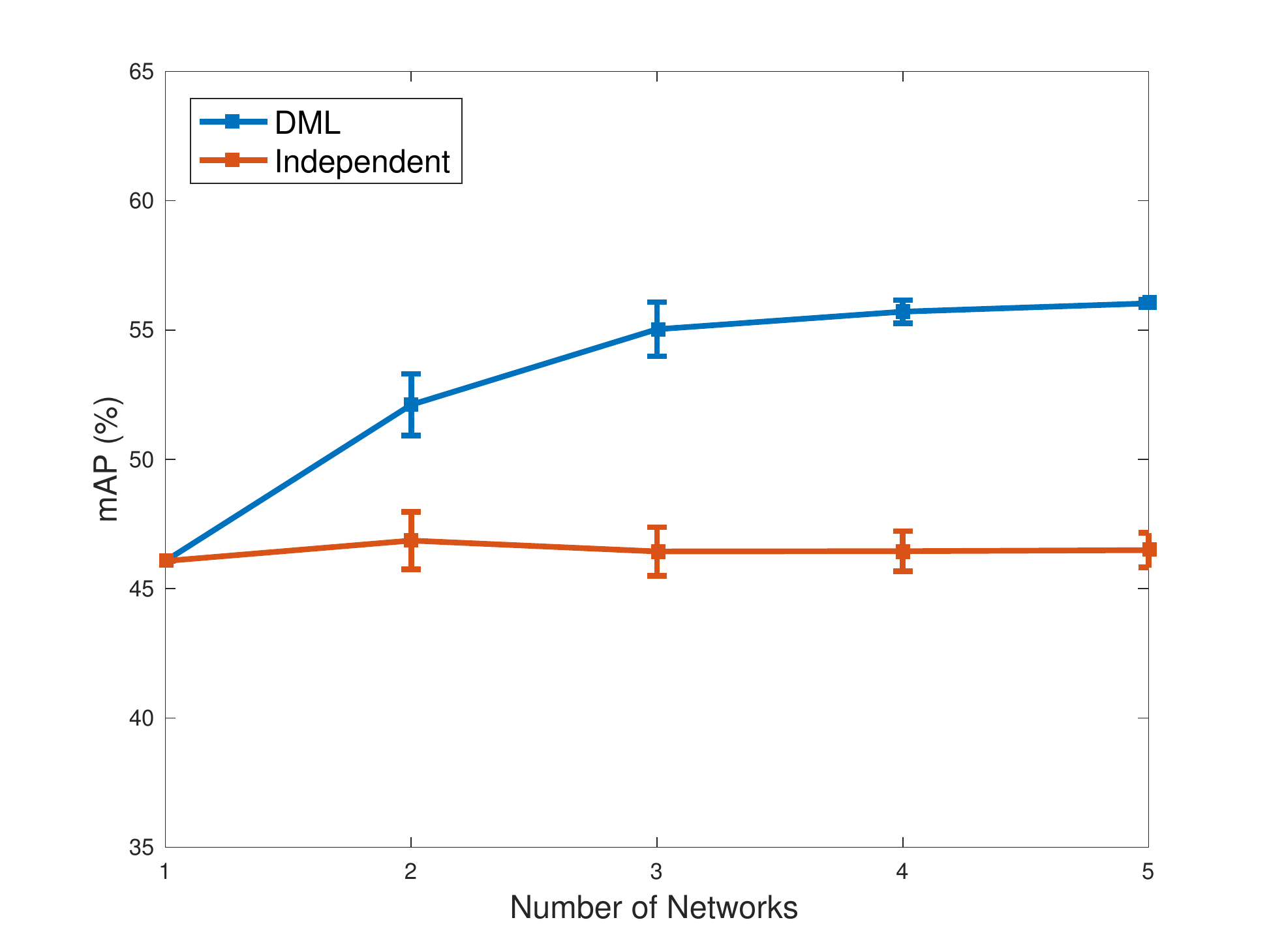} \  & \
   \includegraphics[width=0.47\linewidth, height=0.34\linewidth]{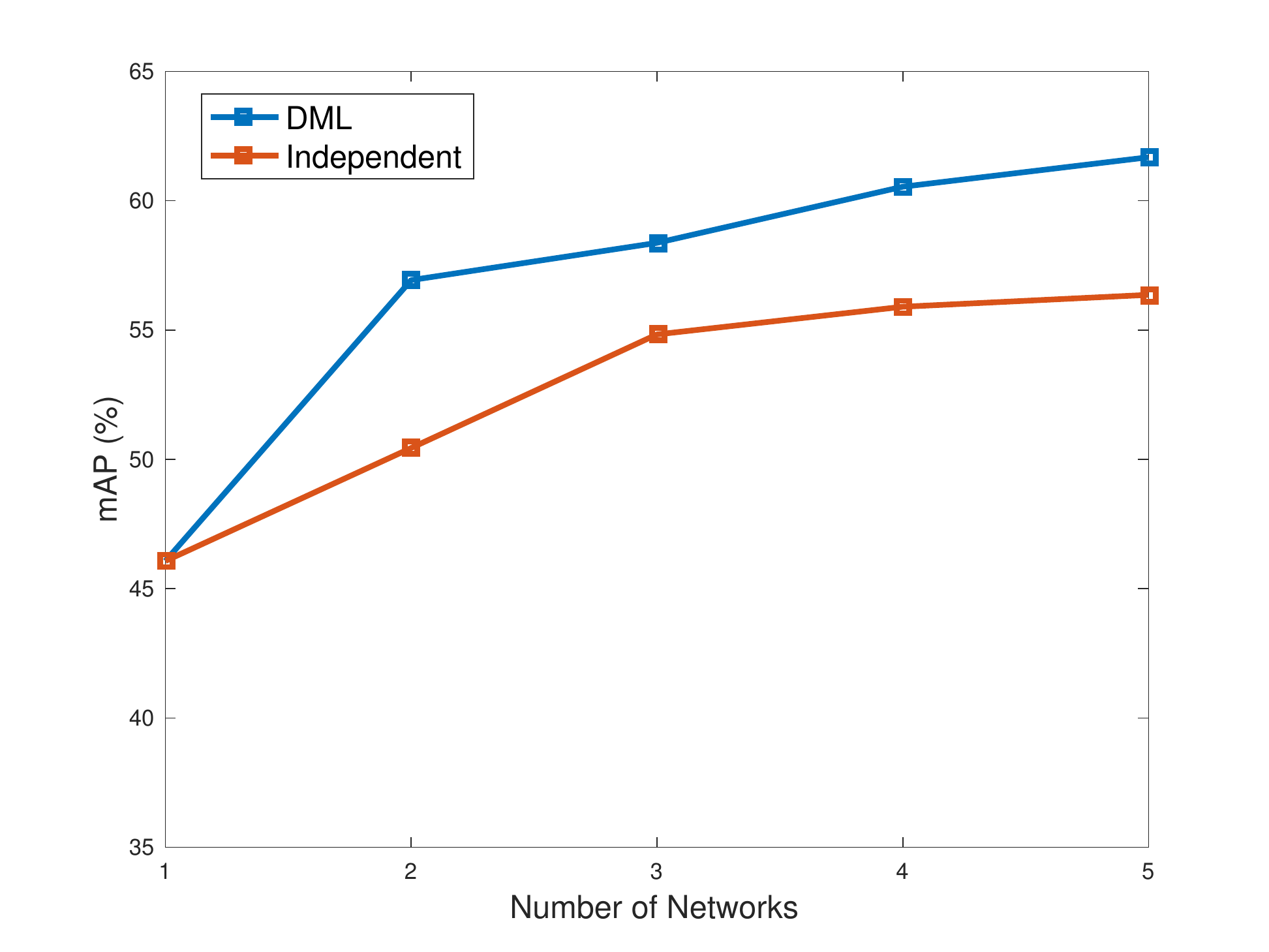}  \\
{\small (a) DML vs.~ Independent} &{\small (b) DML ensemble vs Independent ensemble.} \\
\end{tabular}
  \caption{Performance (mAP (\%)) on Market-1501 with different numbers of networks in cohort} \label{fig:MDML}
\end{figure}

The prior experiments  studied cohorts of 2 students. In this experiment, we study how DML scales with more students in the cohort. Figure~\ref{fig:MDML}(a) shows the results on Market-1501 with DML training of increasing cohort sizes of  MobileNets. The figure shows average mAP, as well as the standard deviation.  From Fig.~\ref{fig:MDML}(a) we can see that the mAP performance of the average \emph{single} network increases with the number of networks in the cohort with DML, hence its gap to the independently trained networks. This demonstrates that the generalisation ability of students is enhanced when learning together with increasing numbers of peers. 
\textcolor{black}{From the standard deviations we can also see that the results get more and more stable with increasing number of networks in DML}. 
 
A common technique when training multiple networks is to group them as an ensemble and make a combined prediction. In Fig.~\ref{fig:MDML}(b) we use the same models as Fig.~~\ref{fig:MDML}(a) but make predictions based on the ensemble (matching based on concatenated feature of all members) instead of reporting the average prediction of each individual. From the results we can see that the ensemble prediction outperforms individual network predictions  as expected (Fig.~\ref{fig:MDML}(b) vs (a)). Moreover,  the ensemble predictions also benefit from training multiple networks as a cohort (Fig.~\ref{fig:MDML}(b) DML ensemble vs.~Independent ensemble).  The ability of DML to improve models ensembles (Fig~\ref{fig:MDML}) illustrates that it may be a generally useful technique to improve performance in applications where model ensembles are standard practice, as there is minimal additional cost if ensembles are already used.
 
\subsection{How and Why does DML Work?}
\label{sec:why DML works}
In this section we attempt to give some insights about how and why our deep mutual learning strategy works. There has been a wave of recent research on the subject of "Why Deep Nets Generalise" \cite{chaudhar2017entropySGD,zhang2017understandingDeepGen,keskar2017batchGen}, which have provided some insights such as: While there are often many solutions (deep network parameter settings) that generate zero train error, some of these generalise better than others due to being in wide valleys rather than narrow crevices \cite{chaudhar2017entropySGD,keskar2017batchGen} -- so that small perturbations do not change the prediction efficacy drastically; and that deep networks are better than might be expected at finding these good solutions \cite{zhang2017understandingDeepGen}, but that the tendency towards finding robust minima can be enhanced by biasing deep nets towards solutions with higher posterior entropy \cite{chaudhar2017entropySGD,pereya2017outputDist}.

\begin{figure}[t]
\centering
  \includegraphics[width=0.999\textwidth, height=0.26\textwidth]{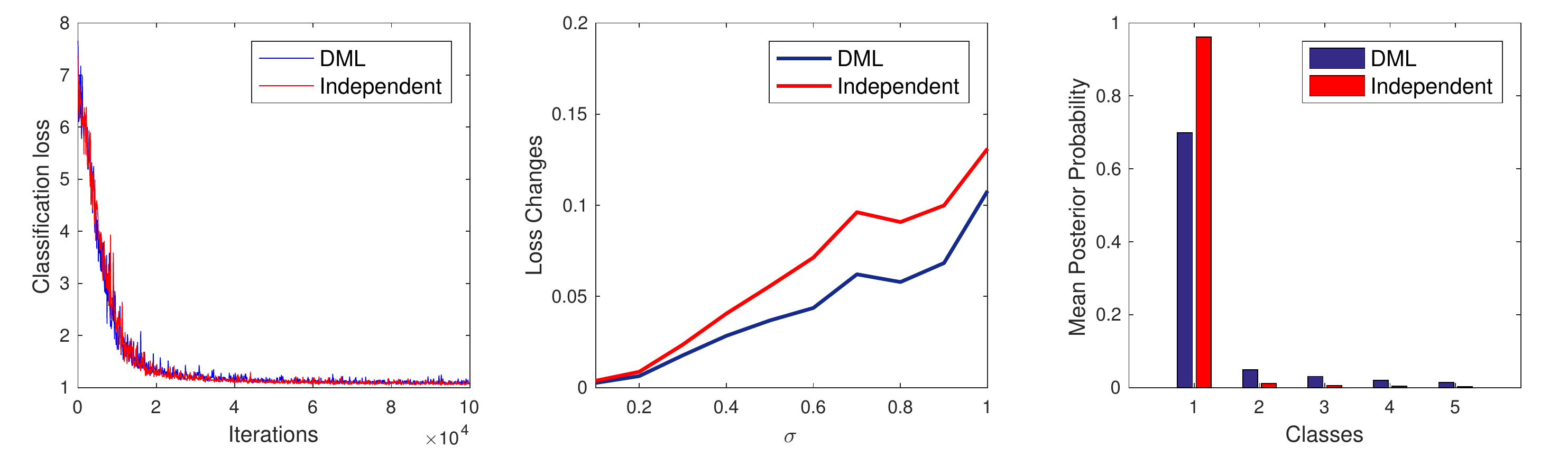}  \\
{\quad \quad \quad \small (a) Training  Loss}  \quad \quad \quad {\small (b) Loss change given parameter noise}  \quad {\small (c) Posterior certainty comparison}
 \caption{Analysis on why DML works}\label{fig:Loss}
 \vspace{-2mm}
\end{figure}

\textbf{Better Quality Solutions with More Robust Minima}\quad With these insights in mind we make some observations about the DML process. Firstly we note that in our applications, the networks fit the training data perfectly: training accuracy goes to 100\%  and classification loss becomes minimal (Fig.~ \ref{fig:Loss}(a)). However, as we saw earlier, DML  performs  better on test data. Therefore rather than helping to find a better (deeper) minima of training loss, DML appears to be helping us to find a wider/more robust minima that generalises better to test data. Inspired by \cite{chaudhar2017entropySGD,keskar2017batchGen}, we perform a simple test to analyse the robustness of the discovered minima on Market-1501 using MobileNet. For the DML and independent models, we compare the training loss of the learned models before and after adding \textcolor{black}{independent Gaussian noise with variable standard deviation  $\sigma$ to each model parameter}. We see that the depths of the two minima were the same  (Fig.~ \ref{fig:Loss}(a)), but after adding this  perturbation the training loss of the independent model jumps up while the loss of the DML model increases much less. This suggests that the DML model has found a much \emph{wider} minima, which is expected to provide better generalisation performance \cite{chaudhar2017entropySGD,pereya2017outputDist}.

\textbf{How a Better Minima is Found}\quad How does  DML  help to find such a better minima? When asking each network to match its peers probability estimates, mismatches where a given network predicts zero and its teacher/peer predicts non-zero are heavily penalised. Therefore the overall effect of DML is that, where each network independently would put a small mass on a small set of secondary probabilities, all networks in the DML tend to aggregate their prediction of secondary probabilities, and both (i) put more mass on the secondary probabilities altogether, and (ii) place non-zero mass on more distinct secondary probabilities. We illustrate this effect by comparing the probabilities assigned to the top-5 highest ranked classes \textcolor{black}{obtained by a ResNet-32 on  CIFAR-100}  trained by DML vs.~an independently trained ResNet-32 model in Fig.~\ref{fig:Loss}(c). For each training sample, the top 5 classes are ranked according to the posterior probabilities produced by the model (Class 1 being the true class and Class 2 the second most probable class,  so on and so forth). Here we can see that the assignment of mass to probabilities below the Top-1 decays much quicker for Independent than DML learning. This can be quantified by the entropy values averaged over all training samples of the DML trained model and the  independently trained model being 1.7099 and 0.2602 respectively. Thus our method has connection to entropy regularisation-based approaches \cite{chaudhar2017entropySGD,pereya2017outputDist} to finding wide minima, but by mutual probability matching on `reasonable' alternatives, rather than a blind high-entropy preference. 

\begin{figure}[h]
  \vspace{-1.5mm}
\centering
  \includegraphics[width=0.92\textwidth, height=0.3\textwidth]{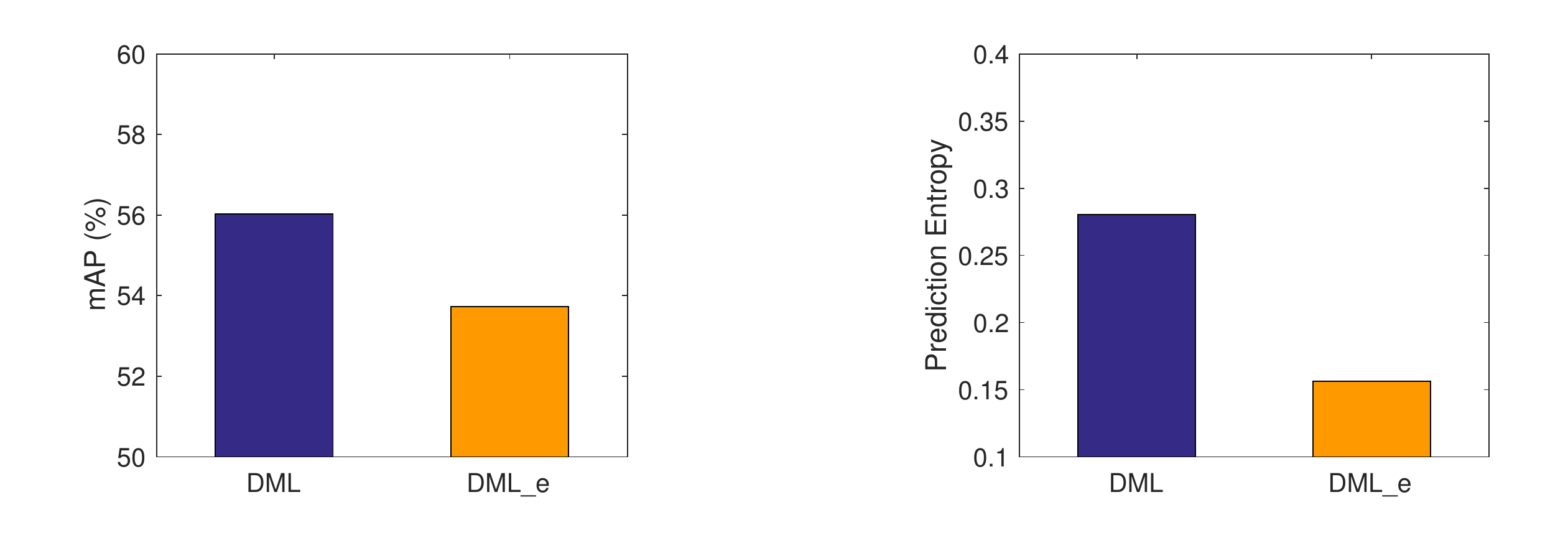}  \\
{\quad \quad \small (a)  Averaged mAP Results   \quad \quad  \quad  \quad  \quad  \quad   \quad  \quad  \quad  \quad  \quad  \quad  \quad \quad   \small  (b) Prediction Entropy } 
 \caption{Comparison of DML with each individual peer student as teacher and DML with peer student ensemble as teacher (DML\_e)  with 5 MobileNets trained on Market-1501 }\label{fig:teahcer}
\vspace{-2mm}
\end{figure}

\textbf{DML with Ensemble Teacher}\quad In our DML strategy, each student is taught by all other students in the cohort individually, regardless how many students are in the cohort (Eq.~(10)). In Sec.~\ref{sec:extension-to-multi-net}, an alternative DML strategy is discussed, by which each student is asked to match the predictions of the ensemble of all other students in the cohort (Eq.~(11)). One might reasonably expect this approach to be better. As the ensemble prediction is better than individual predictions, it should provide a cleaner and stronger teaching signal -- more like conventional distillation. In practice the results of ensemble rather than peer teaching are worse (see Fig.~\ref{fig:teahcer} (a)). By analysing the teaching signal of the ensemble in comparison to peer teaching, the ensemble target is much more sharply peaked on the true label than the peer targets, resulting in larger prediction  entropy value for DML than DML\_e (see Fig.~\ref{fig:teahcer} (b)).  Thus while the noise-averaging property of ensembling is effective for making a correct prediction, it is actually detrimental to providing a teaching signal where the secondary class probabilities are the salient cue in the signal and having high-entropy posterior leads to more robust solutions to model training.

\section{Conclusion}
We have proposed a simple and generally applicable approach to improving the performance of deep neural networks by training them in a cohort with peers and mutual distillation. With this approach we can obtain compact networks that perform  better than those distilled from a strong by static teacher. One application of DML is to obtain compact/fast and effective networks.  We also showed that this approach is also promising to improve the performance of large powerful networks, and that the network cohort trained in this manner can be combined as an ensemble to further improve performance.


{\small
\bibliographystyle{plainnat}
\bibliography{egbib}
}

\end{document}